# Detection of Animal Movement from Weather Radar using Self-Supervised Learning

Mubin Ul Haque[1], Joel Janek Dabrowski[2], Rebecca M. Rogers[3], and Hazel Parry[1]

[1] Agriculture and Food, CSIRO, Dutton Park, Queensland, Australia
{mubin.haque, hazel.parry}@csiro.au
[2] Data61, CSIRO, Dutton Park, Queensland, Australia
joel.dabrowski@data61.csiro.au
[3] Charles Darwin University, Casuarina, Northern Territory, Australia
rebecca.rogers@cdu.edu.au

**Abstract.** Detecting flying animals (e.g., birds, bats, and insects) using weather radar helps gain insights into animal movement and migration patterns, aids in management efforts (such as biosecurity) and enhances our understanding of the ecosystem. The conventional approach to detecting animals in weather radar involves 'thresholding': defining and applying thresholds for the radar variables, based on expert opinion. More recently, Deep Learning approaches have been shown to provide improved performance in detection. However, obtaining sufficient labelled weather radar data for flying animals to build learning-based models is time-consuming and labor-intensive. To address the challenge of data labelling, we propose a self-supervised learning method for detecting animal movement. In our proposed method, we pre-train our model on a large dataset with noisy labels produced by a threshold approach. The key advantage is that the pre-trained dataset size is limited only by the number of radar images available. We then fine-tune the model on a small human-labelled dataset. Our experiments on Australian weather radar data for waterbird segmentation show that the proposed method outperforms the current state-of-the art approach by 43.53% in the dice co-efficient statistic.

**Keywords:** weather radar, flying animals, semantic segmentation, predictive learning, self-supervised learning, aeroecology.

## 1  Introduction

Different flying animals, such as birds, bats, and insects, play crucial roles in our ecosystems as they contribute to pollination, seed dispersal, pest control, and nutrient cycling [1], [2], [3], [4]. In some cases, they may also be considered as pests that are detrimental to an ecosystem, such as an agricultural ecosystem (e.g. plagues of locusts [5]). Therefore, monitoring their movement behavior and migration patterns can help to understand and improve ecosystems.

Weather Radar (WR) was mainly designed to monitor atmospheric conditions, such as rainfall, but it has also been shown to be useful for many decades in detecting flying animals, including birds, bats, insects, and other objects in the airspace [6], [7], [8]. WR emits electromagnetic pulses and measures the reflections of these pulses off objects in the airspace. Given the specific shape and motion of flying animals, the reflections scattered off the animals produce unique signals that can be used to identify the animals. Such reflections are typically referred to as bio-scatter [9], [10].

A common and widely known approach to detect flying animals from WR is to apply a threshold in the weather radar variables. However, this approach is based on interpretation by experts, and thus the process of detecting flying animals' movement using this approach involves a manual and time-consuming process [11], [12], [13], [14].

Given the success of Deep Learning (DL) models to automate many computational tasks with good performance [15], DL models have been applied to automate the task of detecting flying animals' movement from WR images [12], [13], [14], [16]. These DL models are typically based on Supervised Learning (SL) methods, where labelled data is required to identify the WR images with flying animals' movement and used as an invaluable data source to train and evaluate the DL models.

The challenge with SL models is that the models typically require a large labelled WR dataset for training and such datasets are both difficult and time-consuming to obtain [6], [16]. Existing studies that have demonstrated the detection of birds in WR data required substantial manual effort to screen WR images for the presence and absence of flying animals. Fig. 1 shows two examples of WR images with the presence and absence of animals. Manual or expert-driven detection of animal movement from WR images is not scalable due to the large volume of data generated by WR stations [13], [14], [17]. The existing techniques to manually annotate or label WR images include the use of specialized radar (e.g., vertical-looking radar or entomological radar [18]), the use of ecological knowledge of experienced radar ornithology experts [19], the use of citizen science reports [12], and conducting manual survey [20], among others. However, there is a research gap that focuses on building learning-based models with fewer labelled data while detecting animal movement.

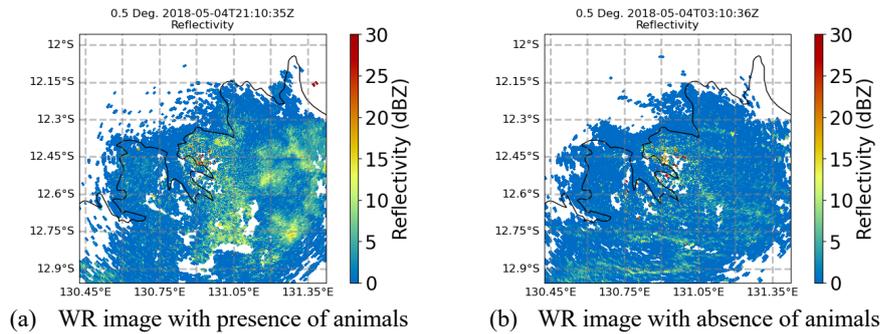

(a)  WR image with presence of animals     (b)  WR image with absence of animals

**Fig. 1** Example of weather radar images with the presence and absence of flying animals captured from Berrimah Weather Radar in Darwin, Australia [20].

In the DL literature, various methods have been proposed to address the challenges associated with the requirements for large datasets, including unsupervised learning [21], semi-supervised learning [22], and self-supervised learning (SSL) [23]. In our SSL methods, a large amount of unlabelled WR data is used to pre-train a neural network model with a threshold approach, without the requirement of any human annotation. With the careful design of a self-supervised problem (which is often referred as a 'pretext' task [23]), the neural network model achieves the ability to capture a high-level representation of input data. Then the neural network model can be fine-tuned with a small amount of labelled data to perform supervised downstream tasks of animal movement segmentation from WR images [24].

In this paper, we propose a novel application of SSL methods for detecting animal movement to tackle the problem of human annotation required for WR images. In brief, we summarize our contributions as follows:

1. To the best of our knowledge, our effort is the first in the aeroecology literature to investigate the applicability of SSL method to detect animal movement in WR images.
2. We demonstrated that ecological expert domain knowledge can be used to create noisy labels for animal movement in WR images, by 'thresholding'. These noisy labels are then used to pre-train our model to distill representative features from the unlabelled data.
3. Our evaluation of the result verified the effectiveness of the incorporation of ecological domain knowledge. Our results indicate that our SSL method significantly outperforms the existing supervised learning method and threshold approach while detecting animal movement from WR images.

This paper is structured as follows: Section 2 describes the related work. The methodology and results are discussed in Section 3 and Section 4, respectively. Finally, the paper is concluded in Section 5.

## 2    Related Work

### 2.1    The threshold approach for detecting flying animals

The threshold approach refers to defining thresholds ('thresholding') from the range of values for weather radar variables that indicate animal movement, based on expert opinion and manual analysis of data in relation to known movement events. For example, Rennie [25] demonstrated that the intensity of the reflectivity factor (Z) was around 10-30 dBZ for locusts. Stepanian et al. [26] demonstrated that the values of the reflectivity factor were around 8-16 dBZ for birds and 0-6 dBZ for insects. Moreover, Gauthreaux and Diehl [9] reported detailed threshold values for different biological scatterers including birds, bats, and insects for different weather radar variables. The analysis of WR data using the threshold approach has helped aeroecology researchers to answer various ecological questions over the years, such as observation of biological scatterers consisting of birds and insects during tropical cyclones [10], population trends in bats [26], [27], and confirmation of birds activities in airspace [11], [29].

### 2.2   Supervised models for detecting flying animals

Chilson et al. [11] made the first attempt to apply learning-based models to study the behavior of flying animals. They applied three different neural network architectures: Artificial Neural Network (ANN), Deep Convolutional Neural Network (DCNN), and a shallow convolutional neural network, to classify radar images that indicated the presence of large aggregations of birds. In particular, they studied Purple Martins (*Progne subis*) and Tree Swallows (*Tachycineta bicolor*). When such birds depart from their roosting sites, they leave a unique 'ring' pattern in the Weather Radar (WR) images. Although the method proposed in [11] successfully classified the WR images as to whether they contain biological scatterers or not, there has been a significant research effort since then to identify the regions (or semantic segmentation) of biological scatterers in WR images.

Cheng et al. [30] proposed a Deep Learning (DL) model based on Faster R-CNN [31] for semantic segmentation to identify bird roosts in WR images. Lin et al. [16] proposed a DCNN based on a Fully Conventional Network (FCN) [32] to discriminate precipitation from biology in WR images. They evaluated deep versus shallow architectures, prediction performances for different elevation angles, and size of training dataset in the model performance [16].

Inspired by the results achieved by the DL models in United States' weather radar [11], [16], [30], Cui et al. [14] and Wang et al. [13] investigated DL models to extract biological scatterers from WR images in China. Cui et al. [14] manually curated 1500 WR images with biological scatterers to train their proposed DL model based on Atrous Separable Convolution [33], whereas Wang et al. [13] manually curated 8000 WR images to train their DL models based on Gated Shape Convolutional Neural Networks (Gated SCNN) [34].

To our knowledge, the most recent development to detect flying animal movement from WR images is proposed by Schekler et al. [12]. Their algorithm was developed with a DL network for semantic segmentation using U-Net architecture [35].

### 2.3   Self-supervised models for detecting flying animals

Although the existing literature [11], [12], [13], [14], [16], [30] have gained significant performance by adopting supervised learning methods, such methods also require a vast amount of labelled annotated data to train the DL model effectively. As we mentioned earlier, the cost and effort to label the WR data remains a major challenge in this domain [23], which can hinder the wider adoption and real-life applicability of the learning models [36]. Self-Supervised Learning (SSL) methods have the potential to address this challenge [15], however, we observed a research gap to investigate the applicability of SSL methods to detect animal movement in WR images.

## 3     Methodology

We introduce a set of symbols to define our segmentation task. For a given weather radar (WR) image, the segmentation task is binary classification of the WR image pixels as belonging to either presence or absence of animal movement [12], [13], [14]. Let $X$ denote a two-dimensional WR image and let $Y$ denote a binary segmentation map which labels the regions in $X$ that have biological activity. The weather radar image dataset ($D$) then comprises two subsets, $D = D_L \cup D_U$, and $D_L \cap D_U = \emptyset$, where $D_L = \{(X_i, Y_i)\}_{i=1}^{N_L}$ contains labels and $D_U = \{(X_i)\}_{i=1}^{N_U}$ has no labels. It is assumed that $N_U \gg N_L$.

Fig. 2 shows a general overview of models that we developed and tested for detecting animal movement from weather radar data. Our proposed self-supervised method consists of the threshold approach, pre-trained model, and fine-tuned model. In self-supervised learning, we leverage unlabelled weather radar data, $D_U$, to pre-train a neural network model $f$ by using the noisy labels obtained from the threshold approach without the need of any human annotation or manual labelling. The model $f$ obtains the ability to capture high-level representation of the unlabelled input data, $\{(X_i)\}_{i=1}^{N_U}$. In fine-tuning, the model $f$ is transferred to supervised downstream tasks by using labelled input data, $\{(X_i, Y_i)\}_{i=1}^{N_L}$. We describe the threshold approach, pre-trained model, fine-tuned model, and supervised model in the following sections.

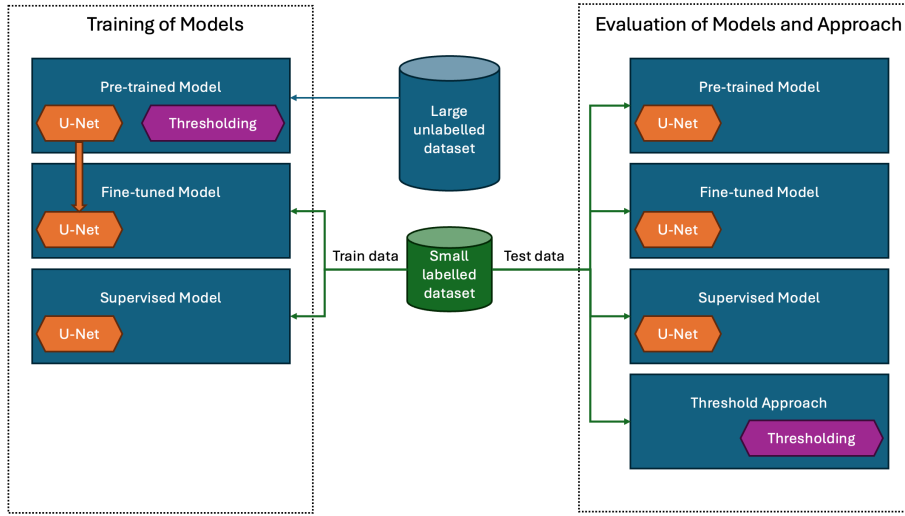

**Fig. 2** A general overview of the four methodologies studied to detect animal movement. Note that the orange arrow connecting the U-Net models in the left panel indicates that the Fine-tuned U-net model uses the Pre-trained U-Net model as a basis for fine-tuning.

### 3.1  Threshold approach

In the threshold approach, a pre-defined threshold is applied to the weather radar variables to identify the presence of biological scatterers, which we describe in detail in Section 2.1. We leverage this threshold approach to generate the noisy labels, $\{(Y'_i)\}_{i=1}^{N_U}$ for unlabelled data $\{(X_i)\}_{i=1}^{N_U}$ to get prior contextual information. The reflectivity factor (Z) was chosen as a range of 8 to 16 dBZ by following the prior research studies [9], [20], [26] for the anticipated presence of animals.

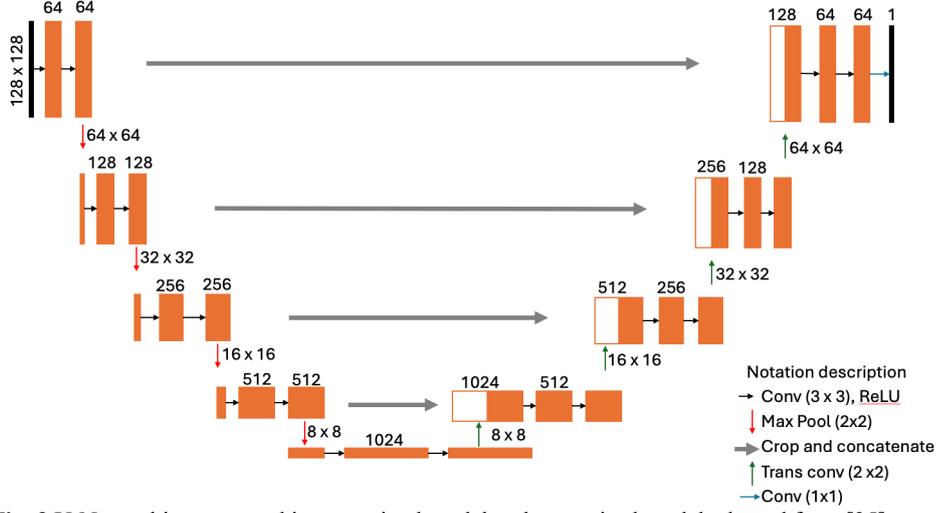

**Fig. 3** U-Net architecture used in pre-trained model and supervised model adopted from [35].

### 3.2  Pre-trained model

After obtaining the noisy labels for the unlabelled data $\{(X_i, Y'_i)\}_{i=1}^{N_U}$, we leverage an U-Net architecture [35] as shown in Fig. 3. The U-Net architecture consists of an encoder (downsampling or contracting block) and follows a regular CNN architecture with convolutional layers, their activations, and pooling layers to downsample the image and extract features from the input image. The encoder consists of the repeated application of two $3 \times 3$ unpadded convolutions, each followed by a rectified linear unit (ReLU) and a $2 \times 2$ max pooling operation with stride 2 for downsampling. The number of feature channels is doubled at each downsampling step.  To create a skip connection, the image is cropped from the contracting path and concatenates it to the current image on the decoding or expanding path. In brief, the height and width of the input images are reduced while the number of channels get increased through several convolutional layers in the encoder or contracting path.

The decoder or expanding path upsamples the feature map, followed by a $2 \times 2$ convolutions (the transposed convolution), where the transposed convolution halves the number of feature channels while increasing the height and width of the image. The next step in the expanding path is a concatenation with the corresponding cropped feature map from the contracting path, and two 3 × 3 convolutions, followed by a ReLU.

In the final layer, a 1 × 1 convolution is used to map each 64-component feature vector to desired number of classes, which is one in the case of biological scatterers extraction from the weather radar images [12], [14]. As a final output from this module, we obtained a pre-trained model $f$ from the unlabelled data $\{(X_i)\}_{i=1}^{N_U}$ and its corresponding noisy label $\{(Y_i')\}_{i=1}^{N_U}$ to learn useful features which were then used to fine-tune the downstream task of animal movement detection from weather radar images (see fine-tuned model, below).

### 3.3   Fine-tuned model

Our self-supervised design incorporates the predictive learning [23] objective, which is predicting a self-produced label $c$, $f(x) \rightarrow c$, which is guided by ecological domain knowledge (the threshold approach). The purpose of the fine-tuned model is to leverage the pre-trained knowledge from the unlabelled data and noisy labelling to improve the performance of the model $f$ on the task-specific labelled data $\{(X_i, Y_i)\}_{i=1}^{N_L}$. The fine-tuning adapts the learned feature representation of the pre-trained model to the specific characteristics of the labelled dataset $D_L$, enhancing the effectiveness of the learning model for the target segmentation task for the detection of animal movement from weather radar.

### 3.4   Supervised model

We also developed a supervised model using the U-Net architecture as shown in Fig. 3. The supervised model uses the task-specific labelled data $\{(X_i, Y_i)\}_{i=1}^{N_L}$.

## 4   Experiment and Results

### 4.1   Dataset

Our proposed method was evaluated on an Australian Weather Radar Dataset for waterbirds [20]. The weather radar station used in [20] to observe the movement of waterbirds is located in Berrimah, Darwin, Northern Territory with geographic coordinates as 12°27′22.3″S 130°55′35.9″E. This weather radar station covers rivers and floodplains which are common habitats for waterbirds [20]. The presence of waterbirds (e.g., when they emerge for feeding flights all at once) was confirmed for 31 days by manual and aerial survey in April and May for 2018 and 2019 [20]. The months of April and May were considered due to the known breeding seasons for waterbirds [37]. The weather radar dataset for 31 days yielded 300 images of 256 × 256 × 3 pixels. 179 images were used to train the supervised model and fine-tune the pre-trained model, whereas 59 and 62 images were used to validate and test the models' performance, respectively. We used the WR dataset of April and May for 2016 and 2017 to prepare the dataset for pre-training. We used the WR dataset of April and May for 2018 and 2019 to train the supervised and fine-tune the pre-trained models. Different year choices ensured that the fine-tuned model was not biased towards the same set of distribution for fine-tuning. We had 1432 weather radar images for pre-training and among them, 859 images were used to train the pre-trained model, whereas 286 and 287 images were used to validate and test the pre-trained model.

### 4.2  Implementation

The weather radar data for Berrimah radar station was downloaded from Australian Unified Radar Archive [38] and pre-processed using pyart [39]. The lowest elevation angle (0.5°) was selected for the weather radar images as it is known to better capture biological scatterer observations compared to higher elevation angles [12], [13], [14]. Moreover, we used the U-Net architecture [35] for all the baseline and proposed approaches, such as pre-trained, supervised, and fine-tuned models so that the comparison among approaches is not affected by the DL architecture choice. Our experiments were carried out in Tensorflow and Nvidia Tesla P100 (SXM2) with 16 GB Memory.

### 4.3  Evaluation Metrics

We used Precision, Recall, and Dice-Coefficient (DC) evaluation metrics to evaluate our proposed method, which are defined as follows.

$$Precision = \frac{TP}{TP + FP} \quad (1)$$

$$Recall = \frac{TP}{TP + FN} \quad (2)$$

$$DC = \frac{2TP}{2TP + FP + FN} \quad (3)$$

Here TP (True Positive) and FN (False Negative) represent the number of positive pixels correctly classified and incorrectly classified respectively, and FP (False Positive) is the number of negative pixels classified incorrectly.

### 4.4  Comparison Study

We compared our proposed fine-tuned model with the threshold approach, pre-trained model, and supervised model as shown in Table 1.

We observed high recall (e.g., 75.03%) and low precision (e.g., 6.98%) for the threshold approach. In addition, we observed similar recall (74.60%) and precision (7.97%) for the pre-trained model which is trained upon the noisy labels generated by the threshold approach. A high recall and low precision suggest that, while the threshold approach and pre-trained model built using the labels from the threshold approach are good at identifying relevant instances (i.e., true positives), these techniques are not specific in segmentation details prediction and tend to generate a high number of false positives.

Fig. 4 (row 4) shows an example of the labels generated by the threshold approach at mid-day (around 12:30 PM), where the appearance of waterbirds is unlikely to be observed in weather radar as the bird activities in aerial space is very minimal due to high temperature at mid-day [40]. From Fig. 4, it is evident that the threshold approach and the pre-trained model tend to be over-sensitive in predicting the animal movement from the weather radar images.

Table 1. Comparison of different models and approaches for detecting animal movement from Australian Weather Radar using the evaluation metrics of precision, recall and dice-coefficient.

| Model and Approaches | Precision (%) | Recall (%) | DC (%) |
| --- | --- | --- | --- |
| Threshold approach | 6.98 | **75.03** | 12.77 |
| Pre-trained Model | 7.97 | 74.60 | 14.39 |
| Supervised Model | **91.47** | 33.94 | 49.50 |
| Fine-tuned Model | 78.04 | 65.21 | **71.05** |

On the other hand, the supervised model is cautious while predicting positive instances, and tends to predict a fewer number of positive predictions overall. When the supervised model does predict a segmentation label, it is highly likely to be correct, which is shown as high precision (91.47%) in Table 1. However, the supervised model misses many positive instances as well, which results in a low recall value (33.94%). We provide an example in Fig. 4 (row 1, time is around 6:00 AM), where the supervised model fails to predict actual animal movement segmentation as positive instances, indicating that supervised model detection is failing to detect some of the initial emergence of waterbirds. In summary, the supervised model reduces the number of false positive generation but is less sensitive in terms of predicting true positive segmentation labels.

The proposed fine-tuned model demonstrates a much better balance between precision and recall, and a significant improvement in the dice coefficient metrics. It achieves a 43.53% improvement over the supervised model [12] for dice-coefficient evaluation metrics and 92.13% for recall evaluation metrics. Fig. 4 shows several examples of the segmentation label prediction by our proposed fine-tuned model with respect to the other models. From Fig. 4, it is exhibited that our fine-tuned model predicts finer segmentation details for detecting animal movement from weather radar.

It is also important to note that, both the supervised and fine-tuned models perform better than the threshold approach in terms of generating fewer false positives and achieving overall performance. For instance, the threshold approach achieves dice co-efficient as 12.77%, whereas the learning-based approaches, such as supervised and fine-tuned models achieve the dice co-efficient as 49.5% and 71.05%, respectively. The high dice co-efficient of our proposed method suggests that our proposed method alleviates the disadvantages associated with the threshold approach, such as, reducing the number of false positive generation. Moreover, supervised and fine-tuned models reduce the clutter (e.g., noise around the radar) in their segmentation prediction while comparing against the threshold approach.

However, note that the fine-tuned model produces less precision value (78.04%) than the supervised model as demonstrated in Table 1. This is expected as the precision and recall are usually not independent and improving one metric tends to reduce another metric and vice versa [41], [42]. Thus, the aim then is to balance these two metrics. In this regard, our proposed method achieved the balanced performance in terms of dice co-efficient. We provide some examples where the fine-tuned model generates a few false positive labelling for animal movement. For instance, from Fig. 4 (row 2 and row 3), some yellow scattered dots are predicted as animal movement, which do not belong to the actual animal movement. One potential approach to reduce such instances of false positives around the animal movement could be the use of the connected component algorithm [43], [44] along with the threshold approach while pre-training the neural network model.

| Weather Radar Image | Ground Truth | Predicted labels by different models | | | |
|---|---|---|---|---|---|
| | | Threshold approach | Pre-trained | Supervised | Fine-tuned |
| 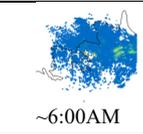 ~6:00AM | 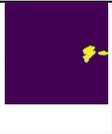 | 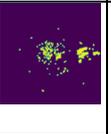 | 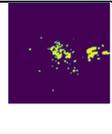 | 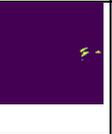 | 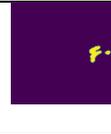 |
| 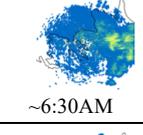 ~6:30AM | 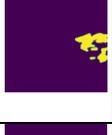 | 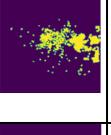 | 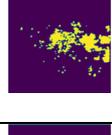 | 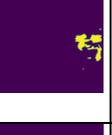 | 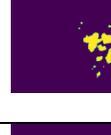 |
| 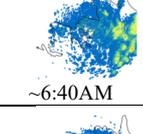 ~6:40AM | 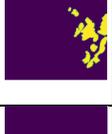 | 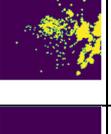 | 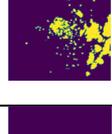 | 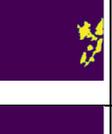 | 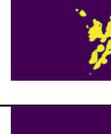 |
| 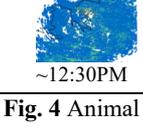 ~12:30PM | 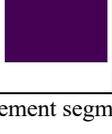 | 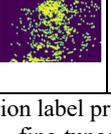 | 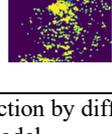 | 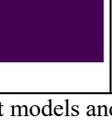 | 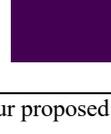 |

**Fig. 4** Animal movement segmentation label prediction by different models and our proposed fine-tuned model.

## 5      Conclusion

 In this paper, we proposed a self-supervised segmentation method based on the threshold approach for animal movement detection from weather radar data. The use of ecological domain knowledge helps to capture prior regions in the weather radar data for animal movement and guide noisy label generation. These noisy labels are used as a data source for the pre-trained model to learn useful feature representation, which is further used for fine-tuning the self-supervised method for animal movement detection. Our model evaluation metrics showed that our method is superior to the latest supervised method as well as the threshold approach. With the rapid generation of a high amount of unlabelled weather radar data, our proposed method has significant potential for identifying animal movement and increasing the effectiveness of segmentation details in weather radar data while reducing the cost and efforts for labelling. In future work, we plan to extend our proposed methods to identify other flying animals, such as bats and insects. We believe that our proposed method will assist ecological professionals and conservation managers in identifying animal movement from weather radar data, which can be used to study their movement patterns and strategic management locations.

**Acknowledgement.** We want to acknowledge Dr Alain Protat and Dr Joshua Soderholm, Bureau of Meteorology, Australia, for their invaluable expertise and insights in processing and analyzing Australian weather radar data.

## References

[1]   S. Hahn, S. Bauer, and F. Liechti, "The natural link between Europe and Africa – 2.1 billion birds on migration," *Oikos*, vol. 118, no. 4, pp. 624–626, 2009, doi: 10.1111/j.1600-0706.2008.17309.x.

[2]   S. Bauer and B. J. Hoye, "Migratory Animals Couple Biodiversity and Ecosystem Functioning Worldwide," *Science*, vol. 344, no. 6179, p. 1242552, Apr. 2014, doi: 10.1126/science.1242552.

[3]   W. Heim *et al.*, "Full annual cycle tracking of a small songbird, the Siberian Rubythroat Calliope calliope, along the East Asian flyway," *J. Ornithol.*, vol. 159, no. 4, pp. 893–899, Oct. 2018, doi: 10.1007/s10336-018-1562-z.

[4]   J. Meade, R. van der Ree, P. M. Stepanian, D. A. Westcott, and J. A. Welbergen, "Using weather radar to monitor the number, timing and directions of flying-foxes emerging from their roosts," *Sci. Rep.*, vol. 9, no. 1, Art. no. 1, Jul. 2019, doi: 10.1038/s41598-019-46549-2.

[5]   D. Lawton *et al.*, "Seeing the locust in the swarm: accounting for spatiotemporal hierarchy improves ecological models of insect populations," *Ecography*, vol. 2022, no. 2, 2022, doi: 10.1111/ecog.05763.

[6]   S. Bauer *et al.*, "The grand challenges of migration ecology that radar aeroecology can help answer," *Ecography*, vol. 42, no. 5, pp. 861–875, 2019, doi: 10.1111/ecog.04083.

[7]   R. M. Rogers, J. J. Buler, C. E. Wainwright, and H. A. Campbell, "Opportunities and challenges in using weather radar for detecting and monitoring flying animals in the Southern Hemisphere," *Austral Ecol.*, vol. 45, no. 1, pp. 127–136, 2020, doi: 10.1111/aec.12823.

[8]   O. Hüppop, M. Ciach, R. Diehl, D. R. Reynolds, P. M. Stepanian, and M. H. M. Menz, "Perspectives and challenges for the use of radar in biological conservation," *Ecography*, vol. 42, no. 5, pp. 912–930, 2019, doi: 10.1111/ecog.04063.

[9]   S. Gauthreaux and R. Diehl, "Discrimination of Biological Scatterers in Polarimetric Weather Radar Data: Opportunities and Challenges," *Remote Sens.*, vol. 12, no. 3, Art. no. 3, Jan. 2020, doi: 10.3390/rs12030545.

[10]  M. S. V. D. Broeke, "Polarimetric Radar Observations of Biological Scatterers in Hurricanes Irene (2011) and Sandy (2012)," *J. Atmospheric Ocean. Technol.*, vol. 30, no. 12, pp. 2754–2767, Dec. 2013, doi: 10.1175/JTECH-D-13-00056.1.

[11]  C. Chilson, K. Avery, A. McGovern, E. Bridge, D. Sheldon, and J. Kelly, "Automated detection of bird roosts using NEXRAD radar data and Convolutional Neural Networks," *Remote Sens. Ecol. Conserv.*, vol. 5, no. 1, pp. 20–32, 2019, doi: 10.1002/rse2.92.

[12]  I. Schekler, T. Nave, I. Shimshoni, and N. Sapir, "Automatic detection of migrating soaring bird flocks using weather radars by deep learning," *Methods Ecol. Evol.*, vol. 14, no. 8, pp. 2084–2094, 2023, doi: 10.1111/2041-210X.14161.

[13]  S. Wang, C. Hu, K. Cui, R. Wang, H. Mao, and D. Wu, "Animal Migration Patterns Extraction Based on Atrous-Gated CNN Deep Learning Model," *Remote Sens.*, vol. 13, no. 24, Art. no. 24, Jan. 2021, doi: 10.3390/rs13244998.

[14]  K. Cui, C. Hu, R. Wang, Y. Sui, H. Mao, and H. Li, "Deep-learning-based extraction of the animal migration patterns from weather radar images," *Sci. China Inf. Sci.*, vol. 63, no. 4, p. 140304, Mar. 2020, doi: 10.1007/s11432-019-2800-0.

[15]  X. X. Zhu *et al.*, "Deep Learning in Remote Sensing: A Comprehensive Review and List of Resources," *IEEE Geosci. Remote Sens. Mag.*, vol. 5, no. 4, pp. 8–36, Dec. 2017, doi: 10.1109/MGRS.2017.2762307.

[16]  T.-Y. Lin *et al.*, "MistNet: Measuring historical bird migration in the US using archived weather radar data and convolutional neural networks," *Methods Ecol. Evol.*, vol. 10, no. 11, pp. 1908–1922, 2019, doi: 10.1111/2041-210X.13280.

[17]  J. Niemi and J. T. Tanttu, "Deep learning–based automatic bird identification system for offshore wind farms," *Wind Energy*, vol. 23, no. 6, pp. 1394–1407, 2020, doi: 10.1002/we.2492.

[18]  V. A. Drake, S. Hatty, C. Symons, and H. Wang, "Insect Monitoring Radar: Maximizing Performance and Utility," *Remote Sens.*, vol. 12, no. 4, Art. no. 4, Jan. 2020, doi: 10.3390/rs12040596.

[19]  H. Mao *et al.*, "Deep-Learning-Based Flying Animals Migration Prediction With Weather Radar Network," *IEEE Trans. Geosci. Remote Sens.*, vol. 61, pp. 1–13, 2023, doi: 10.1109/TGRS.2023.3242315.

Detection of Animal Movement from Weather Radar using Self-Supervised Learning  13

[20] R. M. Rogers, J. Buler, T. Clancy, and H. Campbell, "Repurposing open-source data from weather radars to reduce the costs of aerial waterbird surveys," *Ecol. Solut. Evid.*, vol. 3, no. 3, p. e12148, 2022, doi: 10.1002/2688-8319.12148.

[21] H. U. Dike, Y. Zhou, K. K. Deveerasetty, and Q. Wu, "Unsupervised Learning Based On Artificial Neural Network: A Review," in *2018 IEEE International Conference on Cyborg and Bionic Systems (CBS)*, Oct. 2018, pp. 322–327. doi: 10.1109/CBS.2018.8612259.

[22] X. Yang, Z. Song, I. King, and Z. Xu, "A Survey on Deep Semi-Supervised Learning," *IEEE Trans. Knowl. Data Eng.*, vol. 35, no. 9, pp. 8934–8954, Sep. 2023, doi: 10.1109/TKDE.2022.3220219.

[23] Y. Wang, C. M. Albrecht, N. A. A. Braham, L. Mou, and X. X. Zhu, "Self-supervised Learning in Remote Sensing: A Review." arXiv, Sep. 02, 2022. Accessed: Nov. 28, 2023. [Online]. Available: http://arxiv.org/abs/2206.13188

[24] X. Yang, X. He, Y. Liang, Y. Yang, S. Zhang, and P. Xie, "Transfer Learning or Self-supervised Learning? A Tale of Two Pretraining Paradigms." arXiv, Jun. 19, 2020. doi: 10.48550/arXiv.2007.04234.

[25] S. J. Rennie, "Common orientation and layering of migrating insects in southeastern Australia observed with a Doppler weather radar," *Meteorol. Appl.*, vol. 21, no. 2, pp. 218–229, 2014, doi: 10.1002/met.1378.

[26] P. M. Stepanian, K. G. Horton, V. M. Melnikov, D. S. Zrnić, and S. A. Gauthreaux Jr., "Dual-polarization radar products for biological applications," *Ecosphere*, vol. 7, no. 11, p. e01539, 2016, doi: 10.1002/ecs2.1539.

[27] J. W. Horn and T. H. Kunz, "Analyzing NEXRAD doppler radar images to assess nightly dispersal patterns and population trends in Brazilian free-tailed bats (Tadarida brasiliensis)," *Integr. Comp. Biol.*, vol. 48, no. 1, pp. 24–39, Jul. 2008, doi: 10.1093/icb/icn051.

[28] G. F. McCracken, E. H. Gillam, J. K. Westbrook, Y.-F. Lee, M. L. Jensen, and B. B. Balsley, "Brazilian free-tailed bats (Tadarida brasiliensis: Molossidae, Chiroptera) at high altitude: links to migratory insect populations," *Integr. Comp. Biol.*, vol. 48, no. 1, pp. 107–118, Jul. 2008, doi: 10.1093/icb/icn033.

[29] B. Muller, F. Mosher, C. Herbster, and A. Brickhouse, "Aviation Bird Hazard in NEXRAD Dual Polarization Weather Radar Confirmed by Visual Observations," *Int. J. Aviat. Aeronaut. Aerosp.*, vol. 2, no. 3, Aug. 2015, doi: https://doi.org/10.15394/ijaaa.2015.1045.

[30] Z. Cheng et al., "Detecting and Tracking Communal Bird Roosts in Weather Radar Data," *Proc. AAAI Conf. Artif. Intell.*, vol. 34, no. 01, Art. no. 01, Apr. 2020, doi: 10.1609/aaai.v34i01.5373.

[31] S. Ren, K. He, R. Girshick, and J. Sun, "Faster R-CNN: Towards Real-Time Object Detection with Region Proposal Networks," in *Advances in Neural Information Processing Systems*, Curran Associates, Inc., 2015. Accessed: Dec. 04, 2023. [Online]. Available: https://proceedings.neurips.cc/paper_files/paper/2015/hash/14bfa6bb14875e45bba028a21ed38046-Abstract.html

[32] J. Long, E. Shelhamer, and T. Darrell, "Fully Convolutional Networks for Semantic Segmentation," presented at the Proceedings of the IEEE Conference on Computer Vision and Pattern Recognition, 2015, pp. 3431–3440. Accessed: Dec.

04, 2023. [Online]. Available: https://openaccess.thecvf.com/content_cvpr_2015/html/Long_Fully_Convolutional_Networks_2015_CVPR_paper.html

[33] L.-C. Chen, Y. Zhu, G. Papandreou, F. Schroff, and H. Adam, "Encoder-Decoder with Atrous Separable Convolution for Semantic Image Segmentation," presented at the Proceedings of the European Conference on Computer Vision (ECCV), 2018, pp. 801–818. Accessed: Dec. 04, 2023. [Online]. Available: https://openaccess.thecvf.com/content_ECCV_2018/html/Liang-Chieh_Chen_Encoder-Decoder_with_Atrous_ECCV_2018_paper.html

[34] T. Takikawa, D. Acuna, V. Jampani, and S. Fidler, "Gated-SCNN: Gated Shape CNNs for Semantic Segmentation," presented at the Proceedings of the IEEE/CVF International Conference on Computer Vision, 2019, pp. 5229–5238. Accessed: Dec. 04, 2023. [Online]. Available: https://openaccess.thecvf.com/content_ICCV_2019/html/Takikawa_Gated-SCNN_Gated_Shape_CNNs_for_Semantic_Segmentation_ICCV_2019_paper.html

[35] O. Ronneberger, P. Fischer, and T. Brox, "U-Net: Convolutional Networks for Biomedical Image Segmentation," in *Medical Image Computing and Computer-Assisted Intervention – MICCAI 2015*, N. Navab, J. Hornegger, W. M. Wells, and A. F. Frangi, Eds., in Lecture Notes in Computer Science. Cham: Springer International Publishing, 2015, pp. 234–241. doi: 10.1007/978-3-319-24574-4_28.

[36] I. Rugina, R. Dangovski, M. Veillette, P. Khorrami, and B. Cheung, "Meta-Learning and Self-Supervised Pretraining for Storm Event Imagery Translation".

[37] A. Corriveau *et al.*, "Broad-scale opportunistic movements in the tropical waterbird Anseranas semipalmata: implications for human-wildlife conflicts," *Emu - Austral Ornithol.*, vol. 120, no. 4, pp. 343–354, Oct. 2020, doi: 10.1080/01584197.2020.1857651.

[38] Soderholm J., A. Protat, C. Jakob, "Australian Operational Weather Radar Level 1 Dataset. electronic dataset, National Computing Infrastructure." 2019. doi: 10.25914/508X-9A12.

[39] J. J. Helmus and S. M. Collis, "The Python ARM Radar Toolkit (Py-ART), a library for working with weather radar data in the Python programming language," *J. Open Res. Softw.*, vol. 4, Jul. 2016, doi: 10.5334/jors.119.

[40] B. Rubert, J. O. Branco, G. H. C. Barrilli, D. C. Melo, and A. P. Ferreira, "Behavioral aspects of waterbirds," *Braz. J. Biol.*, vol. 81, pp. 164–177, Mar. 2020, doi: 10.1590/1519-6984.225048.

[41] M. Buckland and F. Gey, "The relationship between Recall and Precision," *J. Am. Soc. Inf. Sci.*, vol. 45, no. 1, pp. 12–19, 1994, doi: 10.1002/(SICI)1097-4571(199401)45:1<12::AID-ASI2>3.0.CO;2-L.

[42] B. Hanczar and M. Nadif, "Controlling and Visualizing the Precision-Recall Tradeoff for External Performance Indices," in *Machine Learning and Knowledge Discovery in Databases*, M. Berlingerio, F. Bonchi, T. Gärtner, N. Hurley, and G. Ifrim, Eds., Cham: Springer International Publishing, 2019, pp. 687–702. doi: 10.1007/978-3-030-10925-7_42.

[43] M. M. Hossam, A. E. Hassanien, and M. Shoman, "3D brain tumor segmentation scheme using K-mean clustering and connected component labeling algorithms," in *2010 10th International Conference on Intelligent Systems Design and Applications*, Nov. 2010, pp. 320–324. doi: 10.1109/ISDA.2010.5687244.

[44] A. Ravankar, Y. Kobayashi, A. Ravankar, and T. Emaru, "A connected component labeling algorithm for sparse Lidar data segmentation," in *2015 6th International Conference on Automation, Robotics and Applications (ICARA)*, Queenstown, New Zealand: IEEE, Feb. 2015, pp. 437–442. doi: 10.1109/ICARA.2015.7081188.